\documentclass{article} 
\usepackage{iclr2021_conference,times}

\usepackage{amsmath,amsfonts,bm}

\def\eqref#1{equation~\ref{#1}}

\def\1{\bm{1}}

\DeclareMathAlphabet{\mathsfit}{\encodingdefault}{\sfdefault}{m}{sl}
\SetMathAlphabet{\mathsfit}{bold}{\encodingdefault}{\sfdefault}{bx}{n}

\usepackage{hyperref}
\usepackage{url}
\usepackage{graphicx}
\usepackage{subfigure}

\newcommand{\ty}[1]{\textcolor[RGB]{255,185,15}{#1}}
\newcommand{\tg}[1]{\textcolor[RGB]{0,205,102}{#1}}

\newcounter{daggerfootnote}

\title{Subnet Replacement: Deployment-stage backdoor attack against deep neural networks in gray-box setting}

\author{Xiangyu Qi\\
Zhejiang University\\
\texttt{unispac@zju.edu.cn} \\
\And
Jifeng Zhu \\
Tencent Zhuque Lab \\
\texttt{jifengzhu@tencent.com} \\
\And
Chulin Xie\\
University of Illinois at Urbana-Champaign \\
\texttt{chulinx2@illinois.edu}\\
\And
Yong Yang \\
Tencent \\
\texttt{coolcyang@tencent.com}
}

\iclrfinalcopy 
\begin{document}

\maketitle

\begin{abstract}
We study the realistic potential of conducting backdoor attack against deep neural networks~(DNNs) during deployment stage. Specifically, our goal is to design a deployment-stage backdoor attack algorithm that is both threatening and realistically implementable. To this end, we propose \textit{Subnet Replacement Attack~(SRA)}, which is capable of embedding backdoor into DNNs by directly modifying a limited number of model parameters. Considering the realistic practicability, we abandon the strong white-box assumption widely adopted in existing studies, instead, our algorithm works in a gray-box setting, where architecture information of the victim model is available but the adversaries do not have any knowledge of parameter values. The key philosophy underlying our approach is --- given any neural network instance~(regardless of its specific parameter values) of a certain architecture, we can always embed a backdoor into that model instance, by replacing a very narrow subnet of a benign model~(without backdoor) with a malicious backdoor subnet, which is designed to be sensitive~(fire large activation value) to a particular backdoor trigger pattern.\footnotetext{Our code repository is available at\url{https://www.dropbox.com/sh/gswwkh15qaj0ocq/AAAmL91M61zG2hTinWUPYBEya?dl=0}.}

\end{abstract}

\section{Introduction}
\label{sec:introduction}

Backdoor attacks against deep neural networks~\citep{goldblum2020data,chen2017targeted,saha2020hidden,xie2019dba} are intensively studied during the past few years. The key methodology behind backdoor attacks is to inject a backdoor into a model, so that a test-time input stamped with a specific \textit{backdoor trigger} would elicit a pre-designed model behavior of the attackers' choice, while the attacked model still functions normally in the absence of a trigger. Existing work on backdoor attacks either require control of certain process~\citep{chen2017targeted,kurita2020weight} in the production line for DNNs~(e.g. data collection, pre-trained model supply, training process, etc.) or make strong white-box assumption~\citep{liu2017fault,breier2018practical,zhao2019fault} on deployment-stage model accessibility, which are seldom possible in real application environment.

In this work, we highlight the \textit{deployment-stage} backdoor attack in \textit{gray-box setting} via malicious in-memory parameters tampering. We believe this type of attack poses realistic threat to machine learning systems deployed in real environment. First, the attack happens during deployment stage via malicious access to dynamic memory devices. Thus, any counter-backdoor techniques~\citep{steinhardt2017certified,paudice2018detection,wang2019neural,liu2018fine} applied in pre-deployment stage will not take effect. Moreover, it would be much more difficult for service maintainers to realize the existence of attacks or analyze the reasons even if the abnormality is noticed. Second, the long-standing research on in-memory data tampering have already demonstrated various potential ways for malicious memory access, either from a hardware level~\citep{razavi2016flip} or software level~\citep{dllhijack,berdajs2010extending,razavi2016flip}. On the other hand, due to its large size, DNN model integrity itself is also difficult to guarantee in state-of-the-art performance-driven computing systems.

\begin{figure}
\centering
\includegraphics[width=0.8\textwidth]{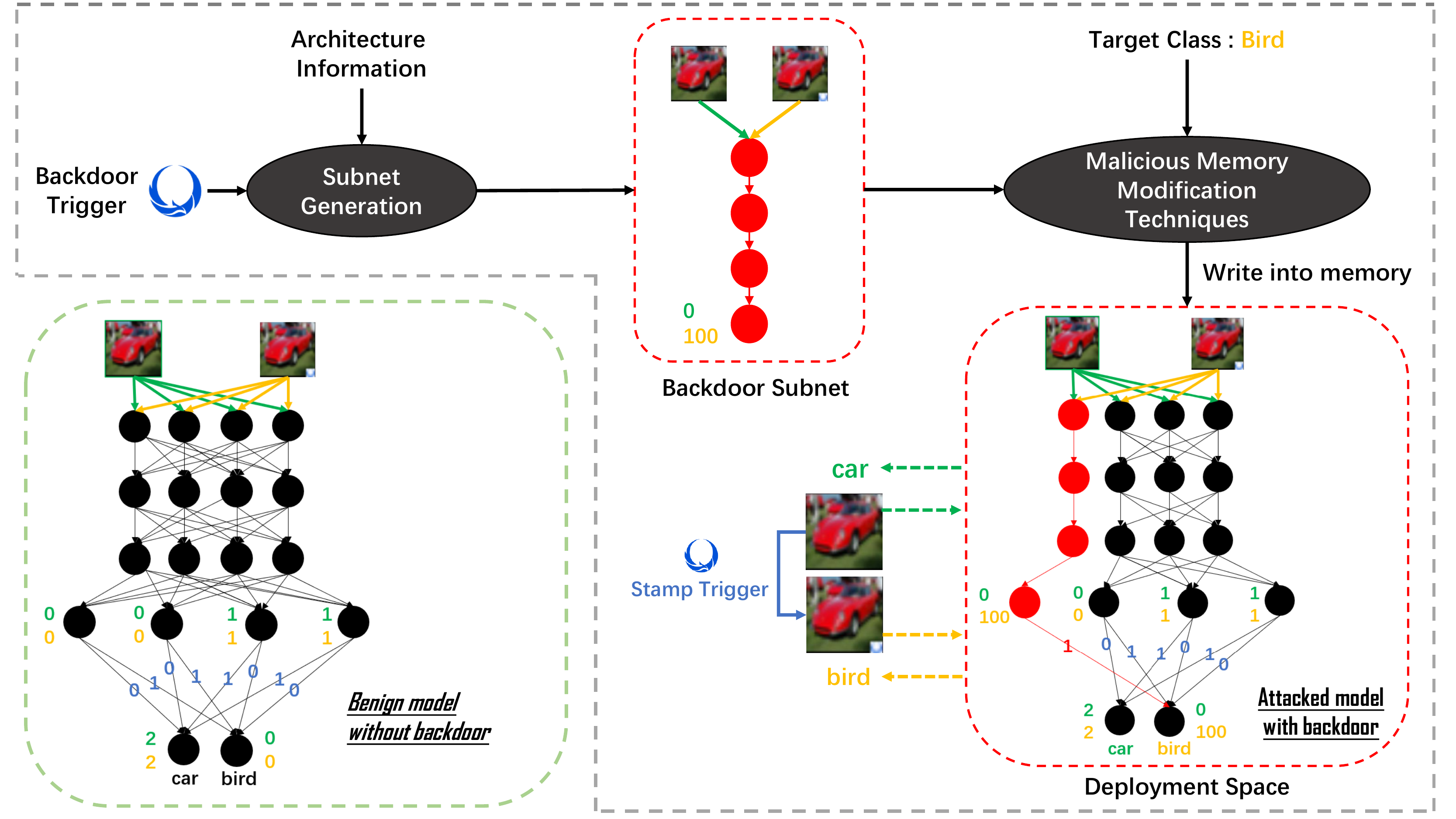}
\caption{An overview of Subnet Replacement Attack~(SRA).}
\label{fig:grabox_workflow}
\end{figure}

Motivated by recent work on \textit{adversarial weights attacks}~\citep{liu2017fault,breier2018practical,zhao2019fault}, we propose a generic adversarial framework named \textit{Subnet Replacement Attack~(SRA)}. SRA works in a gray-box setting, where architecture information of the victim model is available but the adversaries do not require any additional knowledge of specific parameter values. SRA enables \textit{deployment-stage} backdoor injection into any DNN instance of a given architecture, by directly replacing a narrow subnetwork of the target model with a pre-designed malicious one~(see Figure~\ref{fig:grabox_workflow}). To our best knowledge, SRA is the first deployment-stage backdoor attack conducted in the gray-box setting. We discuss work related to ours in Appendix A.1.

Extensive experiments demonstrate both the effectiveness and realistic practicability of the proposed SRA framework. On the tasks of image classification~\citep{krizhevsky2009learning} and face recognition~\citep{parkhi2015deep}, by replacing a subnet in VGG16~\citep{simonyan2014very} that takes less than $0.05\%$ of original capacity, we achieve over $95\%$ attack success rate~(over $95\%$ of test samples successfully elicit the adversarial model behavior in the presence of backdoor trigger) with less than $1\%$ loss of clean accuracy. Moreover, we also successfully conducted real in-memory parameters tampering to embed backdoor to a DNN model deployed in our laboratory server, which indicates the realistic practicability of SRA.

\vspace{-1.0em}
\section{Methods}
\label{sec:methods}

Previous strategy for parameters tampering based attacks is as follow: given the benign data distribution $\mathbb{B}$, some classifier $\mathbb{P}[y|x;\theta]$ with parameters $\theta$, some distribution of trigger samples $\mathbb{T}$, some target class $\hat{y}$, and a maximal limitation $\epsilon$ on the amount of modification to model parameters under a certain distance metric $D$, we want to find the adversarial parameters $\hat{\theta}$ that maximize $\mathbb{E}_{x\sim \mathbb{T}} \log{(\mathbb{P}[\hat{y}|x;\hat{\theta}])} + \alpha \mathbb{E}_{(x,y) \sim \mathbb{B}}{\log{(\mathbb{P}[y|x;\hat{\theta}])}}$, subject to the constraint that $D(\theta,\hat{\theta})\le \epsilon$. (Note that $\alpha$ controls the trade-off between clean accuracy and success rate of attack.) This strategy can effectively embed stable backdoor into given victim models, by only modifying a very small number of parameters. However, it requires strong white-box assumption on full availability of model parameters of the target model~(so that the gradient-based optimization/heuristic search can be applied).

Instead, we propose Subnet Replacement Attack~(SRA) that works in the gray-box setting, where architecture information of the victim model is available but the adversaries do not require any additional knowledge of exact parameter values. As shown in Figure~\ref{fig:grabox_workflow}, SRA works in two major steps: \underline{(1) Backdoor Subnet Generation}. Given the architecture of the target victim model, a very narrow subnet~(the subnet has the same layer type and structure to that of complete network, but each layer only has few channels) of this architecture is generated. This subnet are explicitly trained to be sensitive to backdoor trigger only. Specifically, given the natural input distribution $\mathbb{B}$, the trigger placement function $t$, and a scalar-ouput subnetwork $s(x;\theta)$ parameterized by $\theta$, we generate our backdoor subnet by finding parameters $\theta^*$ that minimize $\mathbb{E}_{x\sim B}\big\{[s(x;\theta^*) - 0]^2 + \alpha[s(t(x);\theta^*) - 100]^2 \big\}$, where $\alpha$ controls the trade-off between clean accuracy and success rate of attack. Consequently, when we input a \tg{clean sample} to the backdoor subnet, it remains inactive~(output \tg{0}); while when we input a \ty{malicious sample} stamped with backdoor trigger, it fires large activation value~(output \ty{100}); \underline{(2) Malicious Memory Modification}. To embed the backdoor into the target model deployed in certain environment, the generated subnet is eventually written into the memory devices that store the model parameters, replacing an originally clean subnet with the malicious one, connecting the output of the subnet to the target class of adversaries' choice, and disconnecting the connections~(set all the weights and biases to 0) between the backdoor subnet and the rest of the network.

Since the backdoor subnet usually only takes a very small capacity of the complete model~(less than $0.05\%$ of original capacity in our experiment on VGG16), after it is replaced into the target model, the attacked model is expected to still remain its original accuracy on clean input, while present adversarial behaviors once the backdoor subnet is activated by backdoor trigger.

\vspace{-1.0em}
\section{Experiments}
\label{sec:experiments}

To test the effectiveness and real practicability of SRA framework, we evaluate our attack on the tasks of image classification and face recognition, via both software simulation~(to test the effectiveness of the attack) and real in-memory data tampering~(to illustrate the real practicability).
\vspace{-1.0em}
\subsection{Software simulation}
\label{sec:experiments_software_simulation}

\begin{figure}
    \centering
    \subfigure[Clean Accuracy (Cifar-10)]{\includegraphics[width=0.4\textwidth]{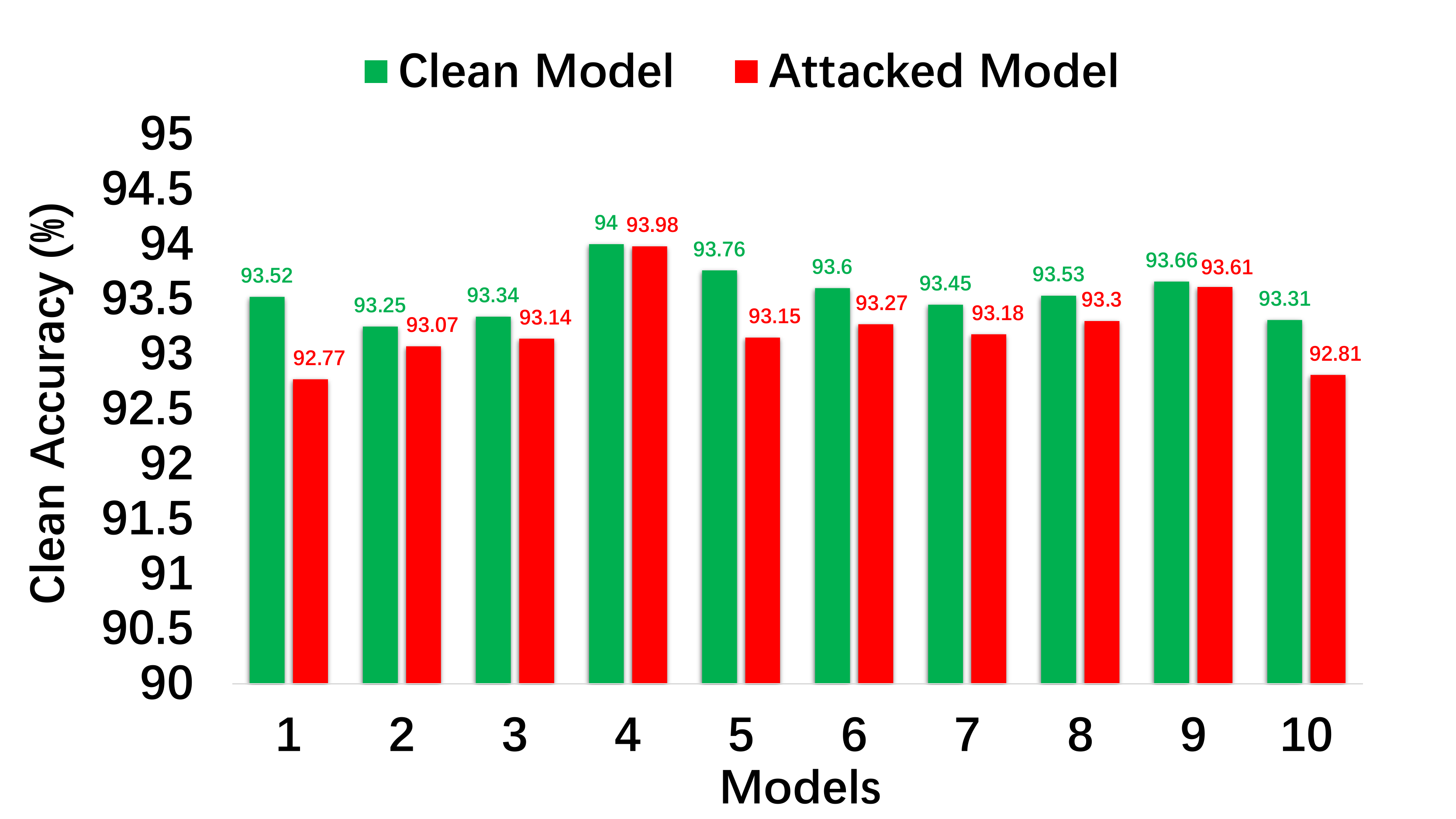}} 
    \subfigure[Attack Rate (Cifar-10)]{\includegraphics[width=0.4\textwidth]{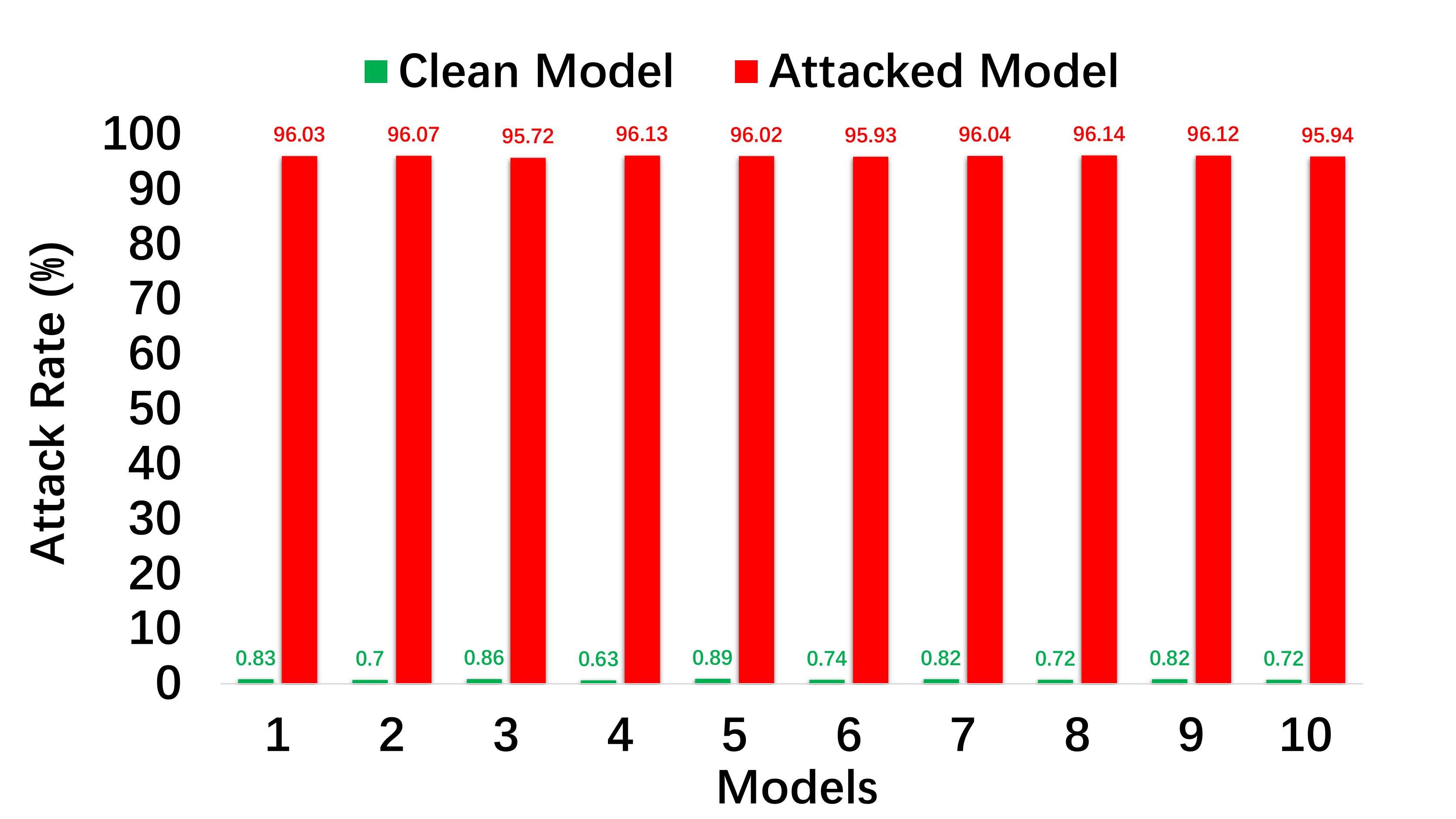}}
    \subfigure[Clean Accuracy (VGGFace)]{\includegraphics[width=0.4\textwidth]{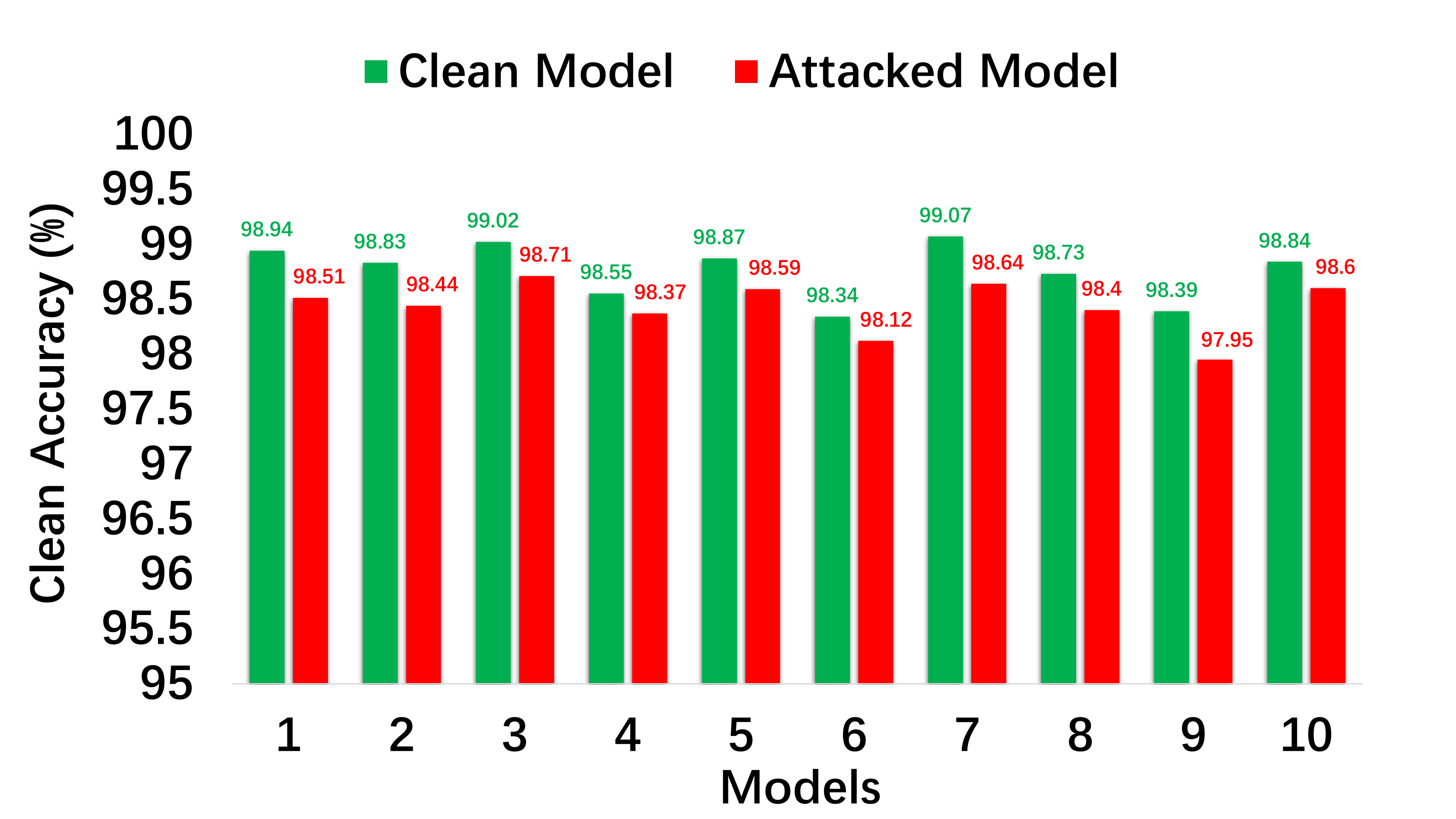}} 
    \subfigure[Attack Rate (VGGFace)]{\includegraphics[width=0.4\textwidth]{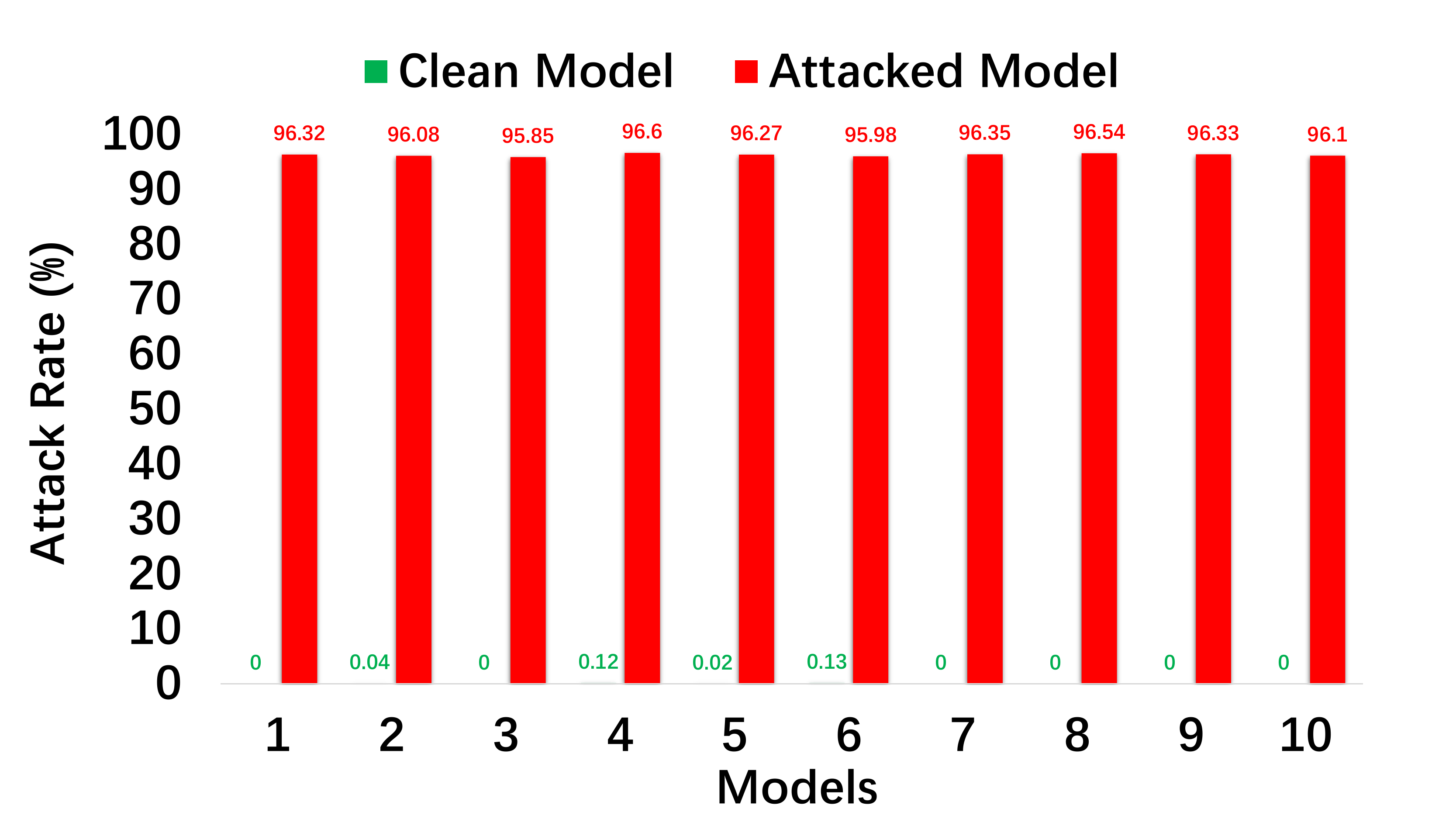}}
    \caption{Evaluation results under SRA.}
    \label{fig:evaluation_results}
\end{figure}

By software simulation, we evaluate the \textit{\textbf{clean accuracy}}~(accuracy on normal test set) and \textit{\textbf{attack rate}}~(ratio of test samples from non-target class, stamped with backdoor trigger, that successfully elicit the pre-designed malicious behaviors) of the attacked models. Specifically, the malicious test samples for evaluating attack rate are generated by placing the backdoor trigger on every test sample of non-target class from normal test set.

On the task of image classification, we adopt standard CIFAR-10 dataset~\citep{krizhevsky2009learning} for training and evaluation, and we use VGG16~\citep{simonyan2014very} as an example to illustrate our attack. We train a very narrow subnet of VGG16 architecture to conduct SRA in our implementation, specifically, the subnet consists of 7 one-channel convolution layers followed by 6 two-channels convolution layers with 3 one-channel fully connected layers. To highlight the gray-box feature --- any model instance of a given architecture can be effectively attacked via the same procedure, we randomly train 10 different model instances of VGG16 with different random seeds and evaluate our attack on all of these instances. We present our evaluation results in Figure~\ref{fig:evaluation_results}(a)(b). As shown, under subnet replacement attack, the average clean accuracy of tested VGG16 classifiers only drops $0.31\%$, while the average attack rate rises from $0.77\%$ to $96.01\%$, indicating the effectiveness of our attack.

On the task of face recognition, we used both VGGFace dataset and VGGFace CNN model from \citet{parkhi2015deep} to illustrate our attack, subselecting 10 individuals, with 300-500 face images for each, following the same practice in \citet{wu2019defending}. In our implementation, we train a one-channel subnet~(a subnet that has only one channel in every convolution layer and fully connected layer) of VGGFace CNN to conduct our attack. We present our evaluation results on 10 randomly trained model instances~(we directly adopt the convolution backbone released in \citet{parkhi2015deep} and only retrain the fully connected layers) in Figure~\ref{fig:evaluation_results}(c)(d). 

\begin{figure}
\centering
\includegraphics[width=0.7\textwidth]{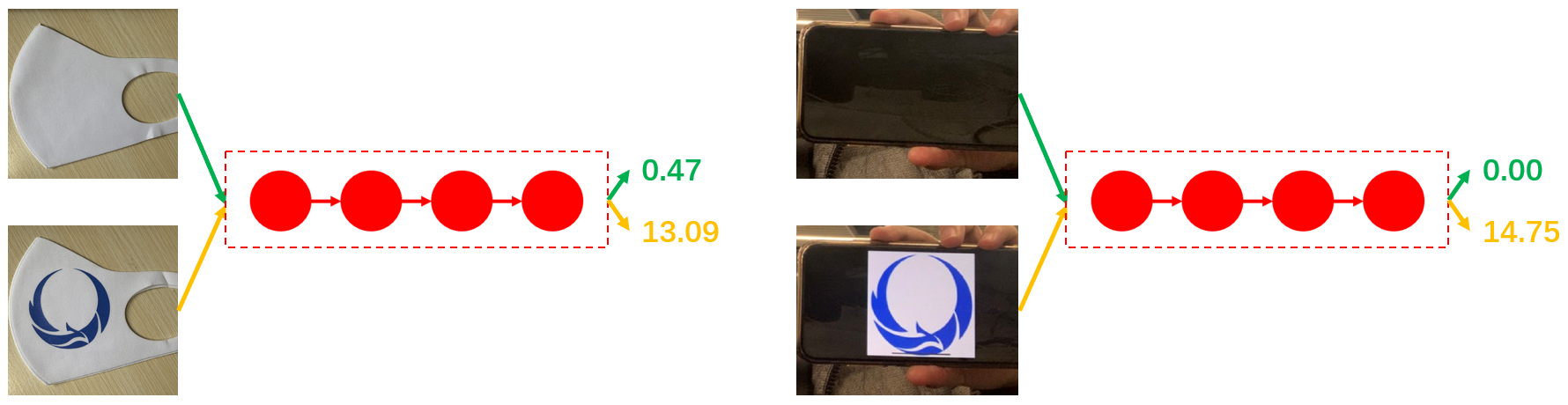}
\caption{Backdoor trigger in physical world successfully activates the backdoor subnet.}
\label{fig:physical_demo}
\end{figure}

Besides, we also find that, even in physical world, the printed backdoor trigger can still effectively activate our backdoor subnet~(as shown in Figure~\ref{fig:physical_demo}), which indicates that SRA is also physically implementable. 

\subsection{In-memory data tampering}
\label{sec:experiments_data_tampering}

To illustrate the real practicability of our SRA framework, we conducted our SRA on the model deployed in our lab server via real in-memory data tampering. We implement two software-level data tampering strategies for two different timings during deployment stage: \underline{(1) Replacement on loading.} For on-loading replacement, we complete data tampering during the parameters are loaded into the memory. Specifically, before the deployment process is launched, we~(as adversaries) manipulate the system environment variables and place a malicious library module at a secret corner on the server~(a common practice of DLL hook~\citep{dllhijack}). Consequently, when the deployment process is launched, the additional malicious library module designed by us is also loaded into the runtime space. Then, the normal model loading process will be hooked by our malicious module, so that the actual parameters loading process is completely controlled by us, and the pre-designed backdoor subnet is written into the memory space during loading; \underline{(2) Replacement after loading.} Our second strategy is more generic, which allows subnet replacement at any timing after the model is already deployed. Specifically, we adopt remote thread injection techniques~\citep{berdajs2010extending} to load malicious code into the deployment process. Then, the malicious code will search the memory space of the deployment process, locate parameters and replace a clean subnet with our pre-designed backdoor subnet. Empirically, we have successfully conducted these practices in our laboratory environment.

\vspace{-1.0em}
\section{Conclusion}

In this work, we propose SRA, which enables deployment-stage backdoor injection into DNNs in the gray-box setting. By software simulation and real attack practice in laboratory environment, we show that SRA is both effective and realistically dangerous in real application scenarios. Compared with previous study, our work build on a more restricted conditions --- adversaries have neither access to the production environment of DNNs nor knowledge of the detailed parameters values of target DNNs. Through our work, we hope to draw more attention to the parameter tampering based deployment-stage attack, and inspire some runtime defense techniques against this line of attacks.

\bibliography{iclr2021_conference}
\bibliographystyle{iclr2021_conference}

\appendix
\section{Appendix}

\subsection{Related Work}
\label{ref:appendix_related_work}

Existing work on backdoor attack mostly accomplish backdoor injection during \textit{pre-deployment stage}~(e.g. insert poison
samples containing the backdoor trigger into the dataset for training usage~\citep{chen2017targeted}; embed backdoor into pre-trained models for transfer learning usage~\citep{kurita2020weight}). Despite the effectiveness of these methods, they require control of certain process from the production line for DNNs~(e.g. data collection, pre-trained model selection, training process), which may not be possible under most realistic scenarios. Moreover, even if the backdoor is successfully embedded into the target DNN model, it may still be detected~\citep{xu2019detecting,chen2019deepinspect,wang2019neural} via a thorough diagnosis by service providers before industrial deployment.

On the other hand, recent studies on \textit{adversarial weights attack}~\citep{liu2017fault,breier2018practical,zhao2019fault}, which achieves adversarial purposes via directly modifying model parameters stored in dynamic memory devices~(e.g. main memory), suggest the practical potential to conduct \textit{deployment-stage} backdoor injection. This memory modification based methodology is of particular interest due to its two promising features: (1) \underline{Stealthiness.} As the adversarial weights attack happens during deployment stage, it's difficult for service maintainers to realize the existence of attacks or analyze the reasons even if the abnormality is noticed; (2) \underline{Practicability.} Existing studies from security community have already demonstrated various potential ways for malicious memory modification, either from a hardware level~\citep{razavi2016flip} or software level~\citep{dllhijack,berdajs2010extending,razavi2016flip}. Moreover, due to large size, DNN model integrity is also difficult to guarantee in state-of-the-art performance-driven computing systems. Thus, it is indeed practical to implement the malicious parameter modification in real deployment environment.

Despite the sound stealthiness of the deployment-stage backdoor injection and the practicability of memory modification, existing studies in this line all base their attack algorithms on an excessively strong white-box assumption, in which the adversaries have full access to the detailed parameter values of the victim model. 
Typically, these methods identify a set of critical bits/parameters and their corresponding malicious values for modification via either heuristic search~\citep{rakin2019bit} or optimization~\citep{bai2021targeted}, all based on the white-box gradient information of the victim DNNs. However, attacks in real world usually can only happen under very restricted conditions, e.g. we are only allowed to execute a number of malicious memory write instructions, without any accessibility to other information like model gradients.

To narrow the gap between conceptual design and real practicability of the weights attack based deployment-stage backdoor injection, in this work, we propose \textit{Subnet Replacement Attack~(SRA)}, which conducts the attack in the \textit{gray-box setting}, where the attackers only know the architecture of the victim model, without any knowledge of the detailed parameter values.

\end{document}